\documentclass{article}

\usepackage{microtype}
\usepackage{graphicx}
\usepackage{subfigure}
\usepackage{booktabs} 
\usepackage{natbib}

\usepackage{hyperref}

\usepackage[accepted]{icml_r2fm}

\usepackage{amsmath}
\usepackage{amssymb}
\usepackage{mathtools}
\usepackage{amsthm}

\usepackage[capitalize,noabbrev]{cleveref}

\theoremstyle{plain}
\newtheorem{theorem}{Theorem}[section]
\newtheorem{proposition}[theorem]{Proposition}
\newtheorem{lemma}[theorem]{Lemma}
\newtheorem{corollary}[theorem]{Corollary}
\theoremstyle{definition}
\newtheorem{definition}[theorem]{Definition}
\newtheorem{assumption}[theorem]{Assumption}
\theoremstyle{remark}
\newtheorem{remark}[theorem]{Remark}

\usepackage[textsize=tiny]{todonotes}

\icmltitlerunning{Don’t Think Twice! Over-Reasoning Impairs Confidence Calibration}

\begin{document}

\twocolumn[

\icmltitle{Don’t Think Twice! Over-Reasoning Impairs Confidence Calibration}





\begin{icmlauthorlist}
\icmlauthor{Romain Lacombe}{yyy}
\icmlauthor{Kerrie Wu}{yyy}
\icmlauthor{Eddie Dilworth}{yyy}

\end{icmlauthorlist}

\icmlaffiliation{yyy}{Stanford University, Stanford, CA 94305, United States}

\icmlcorrespondingauthor{Romain Lacombe}{rlacombe@stanford.edu}

\icmlkeywords{Machine Learning, ICML}

\vskip 0.3in
]



\printAffiliationsAndNotice{}  

\begin{abstract}

Large Language Models deployed as question answering tools require robust calibration to avoid overconfidence. We systematically evaluate how reasoning capabilities and budget affect confidence assessment accuracy, using the \textsc{ClimateX} dataset \cite{climateXpaper} and expanding it to human and planetary health. Our key finding challenges the ``test-time scaling'' paradigm: while recent reasoning LLMs achieve 48.7\% accuracy in assessing expert confidence, increasing reasoning budgets consistently impairs rather than improves calibration. Extended reasoning leads to systematic overconfidence that worsens with longer thinking budgets, producing diminishing and negative returns beyond modest computational investments. Conversely, search-augmented generation dramatically outperforms pure reasoning, achieving 89.3\% accuracy by retrieving relevant evidence. Our results suggest that information access, rather than reasoning depth or inference budget, may be the critical bottleneck for improved confidence calibration of knowledge-intensive tasks.

\end{abstract}

\begin{figure*}[t]
    \centering
    \includegraphics[width=\linewidth]{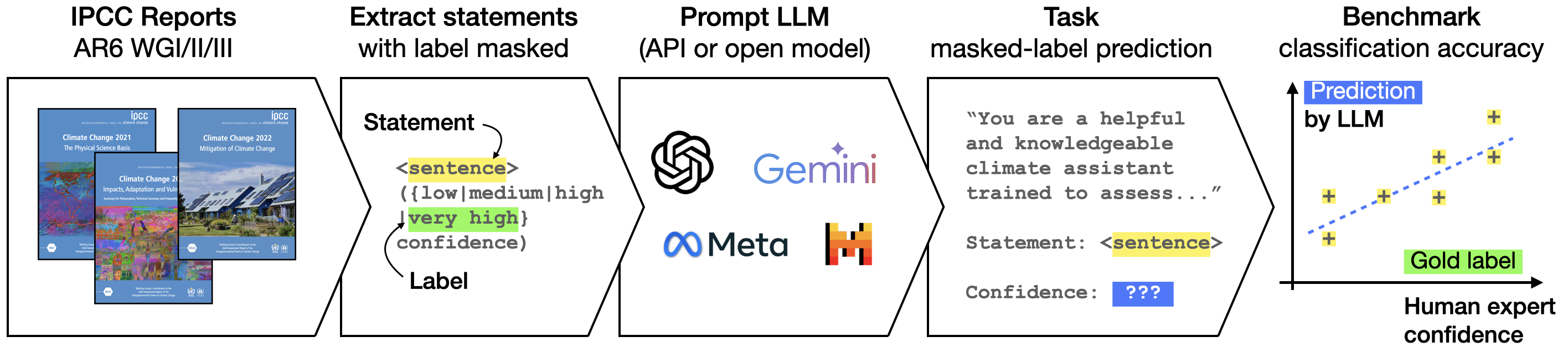}
    \caption{We use the \textsc{ClimateX} dataset \cite{climatex} of statements from the IPCC climate reports \citep{ipcc_wg1, ipcc_wg2, ipcc_wg3} labeled by scientists according to a consistent, categorical confidence scale. We then evaluate and compare LLMs on their ability to reconstruct the ground truth confidence level given a statement with masked label.}
    \label{fig:label-reconstruction-task}
\end{figure*}

\section{Introduction}

The latest generation of Large Language Models (LLMs) exhibits ``reasoning'' abilities, a pattern of inference where models first elaborate long and intricate intermediate chains of thought, which serve as a scratchpad of sorts, before generating their final answer \cite{wei2023chainofthoughtpromptingelicitsreasoning}.
Their widespread adoption, as tools for answering questions and orchestrating agent workflows, calls for careful evaluation of their performance under uncertainty. Calibrating the confidence of these models in particular is notoriously challenging, especially in the absence of objective ground truth as to the accuracy of statements generated in a given domain. 

Accurate calibration is especially important in public-facing domains of science, from climate science to public health, where the large corpora of online text on which LLMs are trained contain long outdated and squarely incorrect content. This is particularly salient as more and more patients turn to AI systems for questions about their health, education, or other high-stakes domains.

Because climate science wrestles with daunting unknowns, from the complexity of the Earth system to the inherent uncertainty of human attempts at mitigating climate change, accurately conveying the level of confidence that experts assign to science and policy statements has long been a central task in the field \cite{Kause2021}. 

This paper builds on the work by climate scientists, who meticulously labeled a vast corpus of climate-related statements with human expert confidence levels, and extends previous work by \citet{climateXpaper} to evaluate the calibration of the latest reasoning models to human expert confidence in statements in the climate domain. 

Specifically, we rely on the \textsc{ClimateX} dataset (Expert Confidence in Climate Statements, \citet{climatex}), a curated, expert-labeled, natural language corpus of 8,094 statements sourced from the 6th Intergovernmental Panel on Climate Change Assessment Report (IPCC AR6) \citep{ipcc_wg1, ipcc_wg2, ipcc_wg3}, and their confidence levels as assessed by scientists based on the quality and quantity of available evidence. 


We use this dataset to study how recent reasoning models compare to the previously reported performance of non-reasoning LLMs on this task \cite{climateXpaper}. Specifically, we ask:

\textbf{(i) Can LLMs accurately assess human expert confidence in climate statements?} We investigate and report experimental results in Table \ref{tab:results}.

\textbf{(ii) Does test-time scaling improve confidence calibration?} We evaluate models with increasing inference budgets, and report results in Figures \ref{fig:accuracy-thinking} and \ref{fig:confidence-thinking}.

\textbf{(iii) Do our results generalize beyond climate?} We introduce a novel dataset in the public health domain, and explore whether reasoning helps or impairs calibration.

\section{Related work}



\subsection{LLMs and linguistic cues of confidence}

Recent literature indicates that expressions of certainty in language influence LLM performance in calibrated NLP tasks. For instance, \citet{Zhou2023} find that appending `weakeners' (i.e., ``A: I think ...'') or `strengtheners' (i.e., ``A: I'm certain...'') to zero-shot question answering prompts can significantly impact LLM performance on common datasets like TriviaQA \cite{joshi2017triviaqa}. Surprisingly, strengtheners led to lower accuracy (40\%) compared to weakeners (47\%) in their experiments, suggesting unique difficulties in reliably interpreting linguistic cues of confidence.

In response to this issue, and to address the broader issue of LLM overconfidence, recent efforts have focused on training models to accurately convey their inherent uncertainty. \citet{lin2022teaching} use supervised fine-tuning \cite{howard2018universal} of a large language model to elicit self-expression of how confidently it answers different arithmetic tasks, using both categorical certainty (e.g., `high confidence') and numeric certainty (e.g., `90\%'). Despite achieving promising outcomes, the authors highlighted potential issues like overfitting to the training data. Calibrating LLMs to linguistic cues of confidence in scientific domains remains a difficult open problem.

\subsection{Assessing what LLMs know}

\citet{kadavath2022language} demonstrated LLMs self-evaluation, showing that models can assess the veracity of their own responses against established human knowledge through few-shot learning. Their work also involved supervised fine-tuning of models to predict the likelihood of accurately answering a given question. Training and evaluation across benchmarks such as \textsc{TriviaQA}, arithmetic, and code generation revealed that while initial self-evaluation performance was low, it significantly improved with few-shot learning, using up to 20 demonstrations.

Previous work has also explored how to evaluate what LLMs ``know.'' \citet{chang2023speak} conducted a `data archaeology' study to deduce which books LLMs were trained on, using a 'name cloze membership inference query' tasks where models are prompted to predict a masked name within a sentence based on its context. Notably, names and sentences are chosen so that human performance on this task is 0\%. The authors identified a strong correlation between the frequency of books in LLM training datasets and the models' performance on the corresponding cloze task.

These results suggest that tasking LLMs to assess human expert confidence by predicting masked labels could be a challenging task, especially in the science domain, for advanced topics uncommon in LLM training sets.

\subsection{LLM evaluation in the climate domain}


Lastly, previous work has assessed how LLMs communicate science information, especially in the climate domain. \citet{bulian2023assessing} introduced a framework for evaluating the fidelity of climate-related information conveyed by LLMs. Their method involved prompting LLMs to generate extensive responses to climate questions, which were then rated by humans across dimensions including accuracy, specificity, completeness, and crucially, the appropriate communication of uncertainty levels. 

IPCC reports have specifically proven to be an excellent context source for LLM systems augmented with retrieval. For instance, ChatClimate \cite{vaghefi2023chatclimate} demonstrated the effectiveness of using scientific content to inform conversational AI agents. Earlier studies have also leveraged climate science literature as a benchmark for NLP systems, such as \textsc{ClimaBench} \cite{laud2023climabench} and \textsc{ClimaText} \cite{varini2021climatext}, which evaluate LLMs on tasks ranging from topic classifications to climate-related QA. Datasets such as \textsc{Climate-Fever} \cite{diggelmann2021climatefever} provide a benchmark to evaluate LLMs on the complex task of verifying climate-related assertions.

\subsection{Reasoning models and test-time scaling}

Much of the recent progress in LLMs has focused on ``test-time scaling,'' the idea that allocating higher compute budgets to generating longer or searching through intermediate chains-of-thought generated by reasoning models could improve their performance \cite{wei2023chainofthoughtpromptingelicitsreasoning}. These techniques have demonstrated vastly improving reasoning abilities, especially in the scientific domain \citep{cui2025curieevaluatingllmsmultitask}, with frontier systems such as OpenAI’s \texttt{o3}, Google's \texttt{Gemini-2.5-Pro}, and Anthropic’s \texttt{Claude 4} achieving accuracies comparable to human experts \citep{openai_o3, gemini_2_5_pro,  anthropic_claude4}.

Models trained to reason demonstrate improved performances in complex domains, especially in sciences \citep{deepseekai2025deepseekr1incentivizingreasoningcapability}. Reinforcement Learning with Verifiable Reward frameworks (RLVR, \citet{lambert2025tulu3pushingfrontiers}), or Reinforcement Fine-Tuning (RFT, \citet{luong2024reftreasoningreinforcedfinetuning}), extends these capabilities by applying policy optimization to LLM roll-outs. Supervised fine-tuning on curated reasoning traces of large models, such as the \textsc{s1k} dataset, also yields reasoning improvements, with scaling laws tying increased reasoning budgets to higher performance on science benchmarks \citep{muennighoff2025s1simpletesttimescaling}.

Test-time scaling is now broadly considered as the next paradigm to improve LLMs, and increased compute is generally expected to translate to direct, measurable gains in task performance. But does increasing the reasoning budget also lead to better calibrated models, with more accurate confidence assessments? We set out to find out.


\section{Dataset}

\subsection{\textsc{ClimateX:} expert confidence in climate statements dataset}

\subsubsection{Human expert confidence in IPCC Assessment Reports statements}

In 2010, the IPCC issued a set of guidelines \cite{Mastrandrea2010} to lead authors of the IPCC Reports on how to consistently communicate uncertainty. Janzwood \cite{Janzwood2020} analyzes the reports written after the guidelines were published and finds evidence of broad adoption across chapter authors and IPCC reports of the `Confidence' framework, which evaluates scientific confidence in each statement by the quality and quantity of available evidence and agreement among peers. 

Confidence is measured on a 5-level categorical scale including `very low,' `low,' `medium,' `high,' and `very high confidence' (with `very low confidence' statements largely excluded from final reports due to lack of evidence). Adoption of this confidence framework by the large scientific community contributing to the IPCC reports makes these Assessment Reports a uniquely rich, deep, and broad corpus of statements in natural language, coherently labeled for confidence, measured as objectively as possible to reflect available evidence and scientific agreement among peers. 


\subsubsection{\textsc{ClimateX} dataset constructions}

This led \citet{climateXpaper} to introduce the \textsc{ClimateX} corpus—the Expert Confidence in Climate Statements dataset—\cite{climatex}, comprising of 8,094 expert-annotated sentences drawn from the three most recent IPCC Assessment Report 6 volumes (Working Groups I, II, and III) \cite{ipcc_wg1, ipcc_wg2, ipcc_wg3}, to evaluate how accurately LLMs assess the confidence level attributed to climate statements by a consensus of human experts.  

The dataset was constructed by extracting the complete raw text from each of the three AR6 IPCC report PDFs using an open-source library \cite{pypdf2}, normalizing the whitespace, tokenizing the text into sentences using NLTK \cite{bird2009natural}, and using regular expression search to filter for complete sentences including a parenthetical confidence label at the end of the statement, of the form: ``\texttt{<statement> (\{low|medium|high|very high\} confidence)}''.

The complete \textsc{ClimateX} dataset contains 8,094 labeled statements, each of them labeled with the corresponding confidence rating on the IPCC’s five-level scale, reflecting the evidence base and the degree of expert agreement. From these sentences, 300 randomly selected statements form a representative test dataset, while the remaining 7,794 statements form the train split. 




\subsection{IARC carcinogenicity expert-labeled dataset}

To expand our analysis outside of the climate domain, we sought another dataset labeled for confidence by a consensus of human experts. The International Agency for Research on Cancer (IARC) reviews and evaluates the evidence on the carcinogenicity of a wide range of human exposures, for purposes of public health and cancer prevention. Based on the strength of available evidence, IARC scientist experts classify agents into one of five levels of confidence regarding their carcinogenic hazard to humans:

\begin{itemize}
    \item Group 1: Carcinogenic to humans. 
    \item Group 2A: Probably carcinogenic. 
    \item Group 2B: Possibly carcinogenic. 
    \item Group 3: Not classifiable. 
    \item Group 4: Probably not carcinogenic. 
\end{itemize}

We extracted the dataset of 1,053 agents and exposure types  classified for carcinogenicity confidence in the IARC Monographs \cite{IARC} since 1971, and used it as the basis for a masked-label classification task benchmark, to evaluate LLMs confidence calibration in the health domain.

The \href{https://huggingface.co/datasets/rlacombe/ClimateX}{\textsc{ClimateX}} and \href{https://huggingface.co/datasets/rlacombe/iarc}{IARC} datasets are available on HuggingFace, and the \href{https://github.com/rlacombe/LLM-Calibration}{source code for experiments} on GitHub.


\section{Experiments}

\subsection{Human expert confidence prediction task}


We compare the performance of recent reasoning and non-reasoning LLMs on the masked label prediction task over the \textsc{ClimateX} dataset, and evaluate them as classifiers against the ground truth labeled by human experts from IPCC reports (see results in Table \ref{tab:results}). 

\textbf{Can LLMs accurately assess human expert confidence in climate statements?} Using the Demonstrate-Search-Predict (DSPy) framework \cite{DSP}, we presented open source LLMs and publicly available APIs for the main commercial models with sentences from the test split of the \textsc{ClimateX} dataset, and instructed them to predict the masked human expert confidence labels. We present our experimental workflow in  Figure \ref{fig:label-reconstruction-task}, and report experimental results in Table \ref{tab:results}.

\textbf{Does test-time scaling improve confidence calibration?} We evaluate models on our benchmark task with increasing inference budgets, and report results in Figures \ref{fig:accuracy-thinking} and \ref{fig:confidence-thinking}.

\textbf{Do our results generalize beyond the climate domain?} We also perform the confidence label reconstruction task on the IARC carcinogenicity dataset, and report results in Figures \ref{fig:accuracy-thinking} and \ref{fig:confidence-thinking}. 




\subsection{Metrics: quantifying confidence levels}





To evaluate model performance, we cast the task as four-way classification over the IPCC confidence categories (“low,” “medium,” “high,” “very high”). We report the following complementary metrics:

\begin{itemize}
\item \textbf{Accuracy} — the proportion of predictions that exactly match the human-assigned label;
\item \textbf{Cohen’s $\kappa$} \cite{cohen1960} — a chance-corrected measurement of inter-rater agreement (here between ground truth and predicted labels), defined as:
\begin{equation}
\kappa = \frac{p_o - p_e}{1 - p_e},
\end{equation}
where $p_o$ is the observed agreement and $p_e$ is the agreement expected under random guessing across the classes (25\% for our four confidence classes).
\item \textbf{Average confidence} — the average predicted score on a dimensionless confidence scale (0.0: no confidence – 3.0: highest confidence).
\end{itemize}

Table \ref{tab:results} presents accuracy and Cohen’s $\kappa$ for every experiment with reasoning and non-reasoning models.

We report 95\% confidence intervals for accuracy and confidence, using bootstrapping with resampling of our experimental results. 

\subsection{Baselines} 

\subsubsection{Non-expert human baseline}
To establish a non-expert human baseline, \citet{climateXpaper} report an experiment with 3 college-educated, non-climate expert participants, who were presented with the statements from the \textsc{ClimateX} test dataset and asked to classify them by their best guess of the consensus confidence level (low, medium, high, or very high). They achieved an accuracy of 36.2\% and Cohen’s 
$\kappa$ of 14.9\% in aligning their estimated confidence assessments with the ground truth provided by climate science experts. This baseline serves as a loose reference point to gauge the performance of LLMs, and highlights the complexity of the task for humans. 
\subsubsection{Classifier baseline}
To establish a baseline for the accuracy achievable through full gradient-descent training, we developed a classifier baseline using the RoBERTa architecture \citep{liu2019roberta}. We fine-tuned the pre-trained \texttt{RoBERTa-Large} model on the \textsc{ClimateX} train set for 2 epochs, with over-sampling of low and very high confidence statements to match the test set distribution. The fine-tuned RoBERTa classifier achieved an accuracy of 53.7\% and Cohen’s $\kappa$ of 38.3\% for prediction of the expert-assigned confidence levels on the test set. 





\begin{table*}[t]
\begin{center}

\begin{tabular}{lcccc}
\textbf{Model} & \textbf{Accuracy} & \textbf{Cohen’s $\kappa$} & \textbf{Bias} & \textbf{Parameters} \\ \hline
 &  &  &  &  \\
Search-Augmented Models &  &  &  &  \\ \hline
\textbf{Google Gemini 2.5 Pro with Search} & \textbf{89.3\%} & \textbf{85.7\%} & +0.030 & Unknown \\
Google Gemini 2.5 Flash with Search & 88.3\% & 84.4\% & +0.097 & Unknown \\
 &  &  &  &  \\
Reasoning Models &  &  &  &  \\ \hline
\textbf{Google Gemini 2.5 Pro} & \textbf{48.7\%} & \textbf{31.6\%} & +0.066 & Unknown \\
Google Gemini 2.5 Pro – Bulk processing & 45.3\% & 27.1\% & +0.353 & Unknown \\
Google Gemini 2.5 Flash – Best thinking budget & 45.0\% & 26.7\% & +0.265 & Unknown \\
OpenAI o3 – Program synthesis & 40.7\% & 20.9\% & +0.167 & Unknown \\
 &  &  &  &  \\
Non-Reasoning Models &  &  &  &  \\ \hline
\textbf{Google Gemini 1.5 Pro} & \textbf{45.0\%} & \textbf{26.7\%} & +0.230 & Unknown \\
OpenAI GPT-4o & 44.0\% & 25.3\% & +0.283 & Unknown \\
OpenAI GPT-4 & 42.4\% & 23.2\% & +0.197 & Unknown \\
OpenAI GPT-3.5 Turbo & 39.7\% & 19.6\% & +0.226 & Unknown \\
 &  &  &  &  \\
Open-Source LLMs &  &  &  &  \\ \hline
\textbf{Meta Llama 3 8B Chat} & \textbf{41.1\%} & \textbf{21.5\%} & -0.001 & 8B \\
Mixtral-8x22B Instruct v0.1 & 38.1\% & 17.1\% & +0.418 & 8$\times$22B \\
Meta Llama 3 70B Chat & 36.2\% & 14.9\% & +0.444 & 70B \\
Mixtral-8x7B Instruct v0.1 & 35.9\% & 14.5\% & +0.303 & 8$\times$7B \\
Mistral 7B Instruct v0.3 & 35.0\% & 13.3\% & +0.423 & 7B \\
Google Gemma Instruct 2B & 33.9\% & 11.9\% & +0.010 & 2B \\
Google Gemma Instruct 7B & 33.4\% & 11.2\% & +0.305 & 7B \\
 &  &  &  &  \\
Baselines &  &  &  &  \\ \hline
\textbf{RoBERTa-Large fine-tuned} & \textbf{53.7\%} & \textbf{38.3\%} & -- & 355M \\
Non-expert humans & 36.2\% & 14.9\% & -- & 60T \\
 &  &  &  & 
\end{tabular}

\end{center}

\caption{Experimental results: accuracy and Cohen's $\kappa$ for the masked confidence label prediction task on \textsc{ClimateX}.}

\label{tab:results}
\end{table*}


\subsection{Experimental controls}

\subsubsection{External validity}
To control for the possibility that LLMs are following our confidence assessment instructions randomly, we conducted experiments using a dataset of statements from The Onion \cite{onion_dataset}, a satirical news website, and grammatically correct but non-sensical sentences from the ReCOGS NLP dataset \cite{wu-etal-2023-recogs}. Models declined to provide an answer at relatively high rates by responding ``I don't know,'' suggesting that labeling results on \textsc{ClimateX} are not simply random predictions.

We also verified external validity of the model prompt on statements outside of the \textsc{ClimateX} dataset. We ran the masked confidence label prediction task on the \textsc{Fever} \cite{Thorne18Fever} True and False sentences, and obtain higher average confidence classification on true statements than on false statements, which serves as a robustness check. 

Lastly, the higher confidence obtained on the IARC carcinogenicity dataset (60-66\%, cf. Figure \ref{fig:confidence-thinking}) suggests that our results generalize beyond the climate domain.

\subsubsection{Self-consistency and robustness to temperature}
To evaluate the self-consistency of LLM answers and their robustness to sampling randomness, we conducted multiple runs of the confidence assessment task on the \textsc{ClimateX} dataset using a temperature of 1. Results for each sentence were largely consistent across runs. For maximum reproducibility, we ran subsequent experiments with a temperature of 0 whenever possible. 

\subsubsection{Robustness to prompt wording}
To investigate whether LLMs react to confidence levels implicitly expressed in the prompt template, we varied the prompts used to elicit confidence predictions from the LLMs, while keeping the underlying \textsc{ClimateX} dataset unchanged. Results were largely unchanged, which confirms that differences between models are a reflection of their capability rather than their prompts.


\begin{figure*}[!t]
    \centering
    \includegraphics[width=1\linewidth]{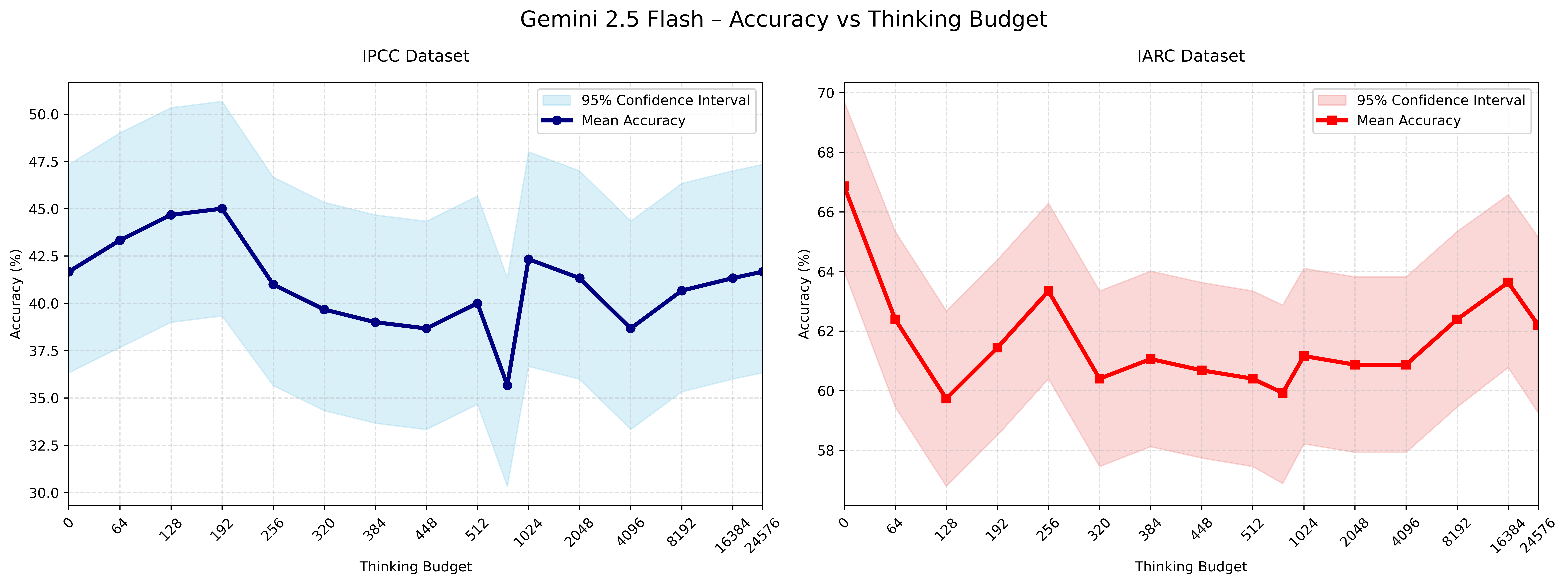}
    \caption{Accuracy (\%) vs thinking budget (tokens) for Gemini 2.5 Flash.}
    \label{fig:accuracy-thinking}
\end{figure*}



\begin{figure*}[!t]
    \centering
    \includegraphics[width=1\linewidth]{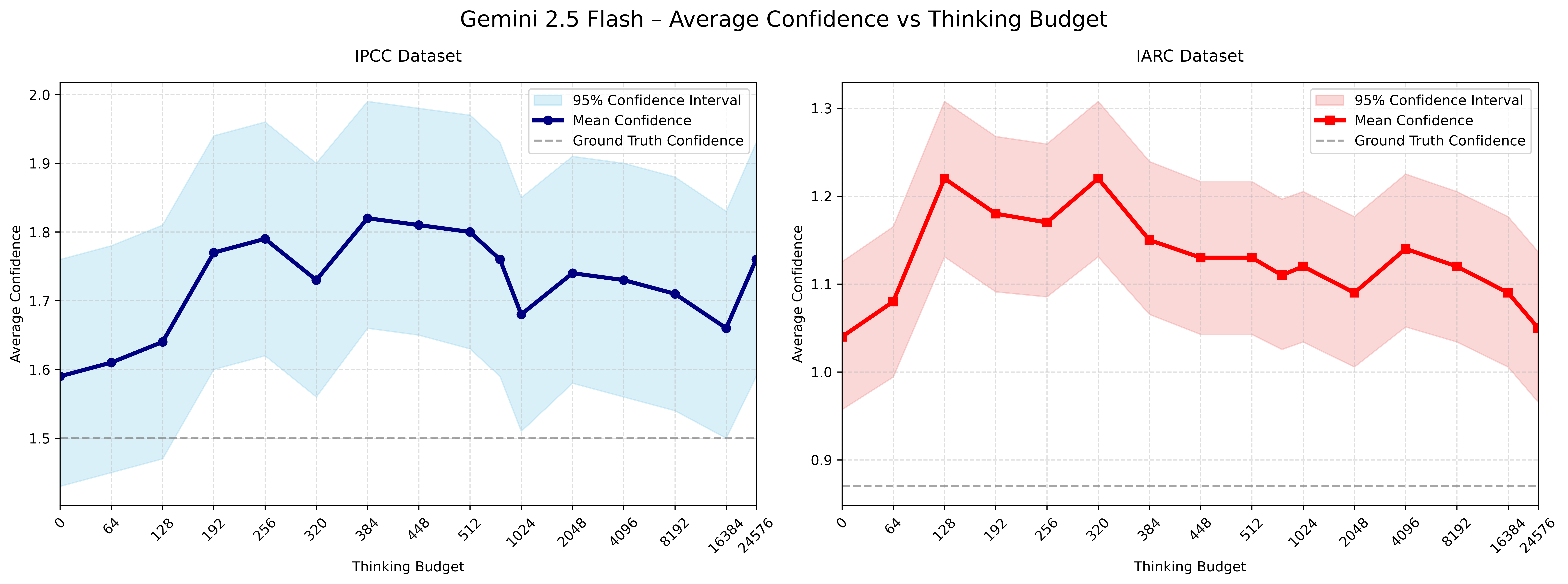}
    \caption{Confidence (dimensionless scale [0.0--3.0]) vs thinking budget (tokens) for Gemini 2.5 Flash.}
    \label{fig:confidence-thinking}
\end{figure*}


\section{Results and analysis}

\subsection{Confidence calibration}

\subsubsection{Recent LLMs can classify human expert confidence in climate-related statements}

Our experiments extend and update results from \citet{climateXpaper} and show how accurately recent LLMs can assess human expert confidence in climate statements: the best performing LLM on our benchmark, \texttt{Gemini-2.5-Pro} can classify masked-label statements in the \textsc{ClimateX} test set with up to 48.7\% accuracy against human expert labels, corresponding to a Cohen's $\kappa$ of 31.6\%. 

Although the confidence prediction task remains challenging for recent LLMs, \textbf{the performance of state-of-the-art models has markedly improved} compared to early 2023 (39.7\% for GPT-3.5), and is undeniably better than the 36.2\% non-expert human baseline. However, it still falls short of the performance of a very small (355M params) encoder model fine-tuned on the \textsc{ClimateX} training set.


Importantly, all models exhibit a positive correlation between ground truth and predicted confidence level, with a slope varying from 0.049 for \texttt{Gemma-Instruct-7B} (near random classifier), up to 0.360 for \texttt{Mixtral-8x22B-Instruct-v0.1}, and classification accuracy that is significantly higher than random chance (25\%).

\subsubsection{Most models are consistently overconfident} 
For a quantitative analysis of confidence calibration, we map IPCC confidence levels to a numerical score, where `low' corresponds to 0, `medium' to 1, `high' to 2, and `very high' to 3. This mapping allows us to treat our LLM classifiers as regression models, and assess their performance by contrasting predicted confidence scores to ground truth numerical values as in a regression setting. 

Meta AI's \texttt{Llama-3-70B-Chat} performs worse on this dimension, consistently making strongly overconfident classifications (+0.444 bias). In contrast with \texttt{Llama-3-8B-Chat}, which exhibits lower bias (close to 0), all other models consistently overestimate certainty level of the statements with which they are presented. Importantly, models specifically overestimate confidence in the `low' and `medium' categories within the \textsc{ClimateX} dataset, which are the most inherently uncertain.

\subsubsection{Size is not destiny} Closed-sourced LLMs are presumed to have large numbers of parameters and to have been pre-trained on vast corpora with vastly larger quantities of compute than currently available open-source models.

Unsurprisingly, the former generally perform better than the latter. The best open-source model (\texttt{Llama-3-8B-Chat}) only out-classed the worst performing closed-source LLM (\texttt{GPT-3.5-Turbo}), which suffers from the additional limitation of a knowledge cutoff date (September 2021) pre-dating the release of the last two IPCC AR6 reports in 2022.

However, among models with known parameter counts, \textbf{model size did not always correlate with improved confidence calibration}. For instance, \texttt{Llama-3-8B-Chat} beats \texttt{LLama-3-70B-Chat}, which exhibits excessive over-confidence, by a large margin; or \texttt{Gemma-Instruct-2B} performs marginally better than the 3.5 times larger \texttt{Gemma-Instruct-7B}.

\subsection{Reasoning and emerging capabilities}

State-of-the-art LLMs offer several orthogonal test-time compute levers to scale performance:

\begin{itemize}
    \item \textbf{Deliberation}: the ability to allocate more internal tokens to ``reasoning'' steps (e.g., \texttt{Gemini 2.5 Flash}’s `thinking budget' feature);
    \item \textbf{Context length}: windows large enough to hold hundreds of statements or even entire corpora and process them at once in bulk.
    \item \textbf{Tool use}: the ability to execute code or call tools such as search engines, to improve completions.
\end{itemize}

We investigate how the first two levers affect confidence calibration, and explore how tool use---specifically Search for retrieval-augmented generation (RAG)---that explicitly injects external evidence in the model context.

\subsubsection{Reasoning and confidence calibration}

\paragraph{Does ``thinking longer'' improve confidence accuracy?} In Figure \ref{fig:accuracy-thinking}, we report the accuracy achieved by \texttt{Gemini 2.5 Flash} for masked label prediction on the \textsc{ClimateX} test set and the IARC dataset against reasoning budget, as we perform a sweep of the thinking budget hyperparameter from 0 to 24,576 (max) tokens.

Three trends emerge:
\begin{itemize}
    \item \textbf{Limited early gains.} A modest budget (64–192 tokens) lifts accuracy from 41.7\% (single-step inference) to a peak of 45.0\% on the IPCC dataset, a +3.3 points jump that mirrors the “chain-of-thought” effect reported by researchers \citep{wei2023chainofthoughtpromptingelicitsreasoning}.
    \item \textbf{Diminishing---and eventually negative---returns.} Beyond ~256 tokens, accuracy collapses, bottoming out near 35.7\% at 768 tokens. The model appears to over-reason, and longer chains of thought may introduce spurious rationales or circular reasoning patterns that hurt both accuracy and calibration. This effect is immediate for the IARC dataset, where introducing reasoning drops performance from 66.9\% to 62.4\%.
    \item \textbf{Partial recovery at extreme budgets.} Pushing past a reasoning budget of 8,192 tokens slowly restores performance, but even the maximum budget ends only slightly above the baseline (41.7\% at 24,576 tokens for IPCC, and 62.2\% for IARC).
\end{itemize}

\paragraph{Does ``thinking longer'' improve confidence calibration?} In Figure \ref{fig:confidence-thinking}, we report the average confidence for masked label prediction by \texttt{Gemini 2.5 Flash} on the \textsc{ClimateX} test set and the IARC dataset against reasoning budget, as we perform a sweep of the thinking budget hyperparameter from 0 to 24,576 (max) tokens. We compute confidence according to the mapping from section 5.1.2, and report the average ground truth confidence level, which an unbiased model should achieve. 

\textbf{Surprisingly, models exhibit increasing and marked over-confidence as the reasoning budget grows}. For instance, \texttt{Gemini 2.5 Flash} grows from +6\% overconfident to +21.3\% overconfident on the IPCC dataset as reasoning budget grows from 0 to 384 tokens, and from +15.6\% to +35.6\% overconfident on the IARC dataset when extending thinking budget to 320 tokens.

Overall, larger reasoning budgets for longer chains of thought appear to impair confidence calibration instead of improving it. These results overall point at a cautionary message for ``test-time scaling'': more inference-time computation alone does not automatically elicit better factuality or improve model confidence calibration to human experts.

\subsubsection{Long context and program synthesis}

The large context windows of recent models enable longer tasks requiring much larger token numbers. This lets us bypass statement-by-statement prompting. We experimented with long context tasks by feeding \texttt{Gemini 2.5 Pro}'s 1M-token window the entire 300-statement test split in a single prompt, asking for one-shot confidence labeling.

The model returned predictions for every line with 45.3\% accuracy---exceeding the statement-by-statement score of \texttt{Gemini 1.5 Pro}---while also producing well-formatted CSV output that worked seamlessly with our evaluation script.

An even more surprising emergent behavior surfaced in \texttt{OpenAI o3}. When supplied the same CSV file, o3 refused to label all 300 statements directly, arguing that the task would be too long. When prompted repeatedly, the model eventually generated a concise Python script to run a smart labeling heuristic based on the presence of certain words (e.g., `high' confidence for sentences including `very likely,' `medium' for `suggest,' `low' for `unlikely'). 

Astonishingly, running the generated script yielded 40.7\% accuracy---lower than the top performing models, but remarkable nonetheless, especially since (i) the model wrote the program unprompted, and (ii) this heuristic exceeds the performance of non-expert humans on the task. 

\textbf{This behavior hints at a form of latent program-synthesis fallback}: when unable to perform the task directly, the model generates an algorithm to solve it programmatically instead. This emerging capability warrants future research and investigation.

\subsubsection{Search-augmented generation}

Finally, we explored retrieval-augmented generation by contextualizing each statement with evidence from an online search performed by the model. This simple RAG setup essentially solved the task close to perfection, and largely saturated our benchmark. With \texttt{Gemini 2.5 Pro} (\texttt{Flash} respectively), search-augmented generation improved accuracy from  48.7\% to 89.3\% (45.0\% to 88.3\%), and Cohen’s $\kappa$ from 31.6\% to 85.7 \% (26.7\% to 84.4\%).

These impressive gains suggest that LLM understanding of complex scientific claims is, to a surprising extent, \textbf{bottlenecked by access to the right evidence rather than by the model’s innate reasoning capacity}. Once the retrieval component surfaces salient passages, the model assigns the correct categorical confidence almost every time. 

These findings echo results from open-book question answering and suggests that, for knowledge-intensive tasks, further scaling of raw model size or ``thinking budget'' may yield diminishing returns relative to investments in retrieval depth and breadth. They also call into question what reasoning capabilities benchmarks actually measure: difficult evaluation sets which allow for Search, such as GPQA \citep{rein2023gpqagraduatelevelgoogleproofqa}, typically do not control for the quality of retrieved evidence, and risk confusing reasoning deficiencies with limitations of knowledge retrieval. 

\subsection{Limitations and future work}



This study acknowledges several limitations that could impact the generalizability of our findings.

\subsubsection{Bias from LLM pre-training data}
Results reported in the IPCC reports are generally well established and therefore extensively published in the literature. As a consequence, they are present in the large corpora of crawled text used for training LLMs, which raises the question of whether benchmark results merely reflect inputs to LLM pre-training, rather than model capabilities. 

This is particularly relevant for fairly comparing model performance---since some models (i.e., \texttt{GPT-3.5-Turbo}) have knowledge cutoff predating the last two IPCC reports, while other models have cutoff dates posterior to all three IPCC reports. Further experiments could use data ablation studies \cite{liu2019inoculation}, or evaluation on held-out unpublished research (e.g., like SWE-Bench \cite{jimenez2024swebench}), to disentangle these effects. 


\subsubsection{Free-form calibration} 
Calibrating models accurately to real-world knowledge and confidence levels is critical for LLMs, with increasingly high stakes as these models are deployed for scientific research, code-generation agents, or information processing workflows with potentially large reach. 

Quantifying how accurately LLM generations match ground truth confidence is an important contribution to address this issue. However, our current approach relies on using the LLM as a classifier rather than in free-form generation. Further work could study how accurately encoder-based models, such as the classifier we trained as our baseline, rate the accuracy of LLMs confidence in longer outputs.

\subsubsection{Evolving confidence baselines}
The human expert confidence labels we use as our ground truth are inherently a moving target, formed through consensus based on available evidence at a given time. Future work could study how these confidence assessments evolve over time as new literature and evidence emerge. 

The dynamic nature of confidence levels underscores the need for ongoing evaluation and adaptation of NLP models and benchmarks \cite{kiela2021dynabench}. Notably, AI for Science would benefit from a better understanding of how to update implicit confidence levels on the basis of new experimental results.

 \subsubsection{Limited human baselines} The human baseline reported by \citet{climateXpaper} was limited to three non-expert participants, and may be biased as it was performed while finalizing the \textsc{ClimateX} test set. Future work could include an updated baseline with more participants for better statistical representativeness, and include the IARC dataset. 
 
 The baseline also leaves out the question of how expert humans themselves perform on this task. Given the complexity and the multidisciplinary nature of climate sciences, it is possible that even domain experts would not perform well on our benchmark. Better understanding the limits of human expert performance would inform future developments of new LLM evaluations building on \textsc{ClimateX}, and future work could include a baseline study with domain experts, such as authors and chapter leads of the IPCC reports.



\section{Conclusion}


This study fundamentally challenges the assumption that more reasoning automatically leads to better-calibrated AI systems. Our systematic evaluation using expert-labeled climate science statements reveals that extended reasoning consistently impairs rather than improves confidence calibration, leading to systematic overconfidence that worsens with longer thinking budgets.

Calibrating the frequently ``confidently wrong'' LLMs is critical for language model-based applications to function effectively as knowledge retrieval systems. The ability to benchmark whether generated outputs adequately convey the underlying ground truth confidence is paramount for LLMs to meet that objective.


This is especially true for reasoning models, which are now often forming the basis for multi-agent systems and modular architectures that rely on the accuracy of the underlying model to perform complex tasks. Evaluating, measuring, and improving their confidence calibration is therefore critical for the fields of reasoning agents, large language models, and Machine Learning systems.

In this work, we've explored whether ``test-time scaling'' and increased inference time compute budgets help the latest generation of reasoning models improve their factuality and their confidence calibration. \textbf{The answer is ``no'': while our benchmark is topped by the reasoning models, increasing their thinking budget makes models overconfident and impairs their confidence calibration!}


To reach these conclusions, we expanded on previous work and used the \textsc{ClimateX} dataset, a curated, natural language corpus of 8,094 statements sourced from the 6th IPCC Assessment Report, expert-labeled for confidence levels by climate scientists. We also expended our confidence calibration benchmark to public health, epidemiology, oncology, and toxicology, with the IARC dataset of agent exposure carcinogenicity confidence.

To our knowledge, ours is the first study to systematically document calibration of LLMs to human expert confidence levels in reasoning models, and to investigate how reasoning budget impacts confidence calibration in the human and planetary health domain. 

Still, a key result we report is a negative one: we show that increasing reasoning budget through ``test-time scaling'' impairs the confidence calibration of reasoning models. 

Our finding that search-augmented generation dramatically outperforms even the best reasoning models (89.3\% vs 48.7\% accuracy) points toward a more promising path forward. Rather than scaling reasoning compute, practitioners should prioritize improving information retrieval and evidence synthesis capabilities. This aligns with broader trends in AI research suggesting that access to relevant information often matters more than raw computational power for knowledge-intensive tasks.

\section*{Acknowledgments}

The authors wish to thank Christopher Potts and Mina Lee. We are also grateful to the thousands of scientists and experts who contributed to the IPCC AR6 reports and the IARC monographs, and the many more whose research informed them; this work would not have been possible without them.





\bibliography{icml_r2fm_refs}
\bibliographystyle{icml2025}

\appendix


\end{document}